\documentclass[twoside]{article}
\usepackage{nips12submit_e,times}
\usepackage{amsmath,amssymb,amsthm,amsfonts}
\usepackage{graphicx}
\usepackage{multirow}
\usepackage{sidecap}
\usepackage{multicol}
\usepackage{subfigure}
\usepackage{wrapfig}
\usepackage[numbers,sort&compress]{natbib}
\usepackage{algorithmic}
\usepackage{algorithm}

\newcommand{\Sp}{\ \ }

\title{Automated  Variational Inference \\in Probabilistic Programming}
\author{David Wingate, Theo Weber}

\nipsfinalcopy 
\begin{document}
\maketitle

%
%

\begin{abstract}

We present a new algorithm for approximate inference in probabilistic programs, based on a stochastic gradient for variational programs. This method is efficient without restrictions on the probabilistic program; it is particularly practical for distributions which are not analytically tractable, including highly structured distributions that arise in probabilistic programs.  We show how to automatically derive mean-field probabilistic programs and optimize them, and demonstrate that our perspective improves inference efficiency over other algorithms.

\end{abstract}

%
%

\section{Introduction}

\subsection{Automated Variational Inference for Probabilistic Programming}

Probabilistic programming languages simplify the development of probabilistic models by allowing programmers to specify a stochastic process using syntax that resembles modern programming languages. These languages allow programmers to freely mix deterministic and stochastic elements, resulting in tremendous modeling flexibility.  The resulting programs define prior distributions: running the (unconditional) program forward many times results in a distribution over execution traces, with each trace generating a sample of data from the prior.  The goal of inference in such programs is to reason about the posterior distribution over execution traces conditioned on a particular program output.  Examples include BLOG \citep{milch05}, PRISM \citep{sato97}, Bayesian Logic Programs \citep{kersting07}, Stochastic Logic Programs \citep{Muggleton96stochasticlogic}, Independent Choice Logic \citep{poole08}, IBAL \citep{pfeffer01}, Probabilistic Scheme \citep{radul07}, $\Lambda_\circ$ \citep{park08}, Church \citep{goodman08}, Stochastic Matlab \citep{wingate11}, and HANSEI \citep{kiselyovS09}.

It is easy to sample from the prior $p(x)$ defined by
a probabilistic program: simply run the program.  But inference in such languages is hard:
given a known value of a subset $y$ of the variables, inference must
essentially run the program `backwards' to sample from $p(x|y)$.
Probabilistic programming environments simplify inference by providing
universal inference algorithms; these are usually sample based (MCMC
or Gibbs) \citep{goodman08,milch05,pfeffer01}, due to their
universality and ease of implementation.

Variational inference~\citep{beal2003variational,jordan99,winn2006variational}  offers a powerful, deterministic approximation
to exact Bayesian inference in complex distributions.  The goal is to
approximate a complex distribution $p$ by a simpler parametric
distribution $q_\theta$; inference therefore becomes the task of
finding the $q_\theta$ closest to $p$, as measured by KL divergence.
If $q_\theta$ is an easy distribution, this optimization can often be
done tractably; for example, the mean-field approximation assumes that
$q_\theta$ is a product of marginal distributions, which is easy to
work with.

Since variational inference techniques offer a compelling alternative to sample
based methods, it is of interest to automatically derive them,
especially for complex models. Unfortunately, this is intractable for
most programs.  Even for models which have closed-form coordinate
descent equations, the derivations are often complex and cannot be
done by a computer.  However, in this paper, we show that it \emph{is} tractable to
construct a stochastic, gradient-based variational inference algorithm
automatically by leveraging compositionality.

\section{Automated Variational Inference}
\label{sec:rl}

An unconditional probabilistic program $f$ is defined as a
parameterless function with an arbitrary mix of deterministic and
stochastic elements. Stochastic elements can either belong to a fixed
set of known, atomic random procedures called ERPs (for
\emph{elementary random procedures}), or be defined as function of
other stochastic elements.

The syntax and evaluation of a valid program, as well as the
definition of the library of ERPs, define the probabilistic
programming language. As the program $f$ runs, it will encounter a
sequence of ERPs $x_1,\cdots,x_T$, and sample values for each.  The
set of sampled values is called the \emph{trace} of the program. Let
$x_t$ be the value taken by the $t^{\text{th}}$ ERP.  The probability
of a trace is given by the probability of each ERP taking the
particular value observed in the trace:
\begin{equation}
\label{eq:pplik}
p(x) = \prod_{t=1}^T p_t (x_t\:|\:\psi_t(h_t))
\end{equation}
where $h_t$ is the history $(x_1,\ldots,x_{t-1})$ of the program up to
ERP $t$, $p_t$ is the probability distribution of the $t^{\text{th}}$
ERP, with history-dependent parameter $\psi_t(h_t)$.  Trace-based
probabilistic programs therefore define directed graphical models,
but in a more general way than many classes of models, since the
language can allow complex programming concepts such as flow control,
recursion, external libraries, data structures, etc.

\subsection{KL Divergence}

The goal of variational inference \citep{jordan99,beal2003variational,winn2006variational} is to approximate
the complex distribution $p(x|y)$ with a simpler distribution
$p_{\theta}( x )$.  This is done by adjusting the parameters $\theta$
of $p_{\theta}( x )$ in order to maximize a reward function $L(\theta)$,
typically given by the KL divergence:
\begin{eqnarray}
\label{eq:kl}
KL(p_\theta,p(x|y)) & = & \int_x p_{\theta}( x ) \log \left(
  \frac{p_\theta(x)}{p(x|y)} \right) \nonumber \\
& = & \int_x p_{\theta}( x ) \log \left(
  \frac{p_\theta(x)}{p(y|x)p(x)} \right) + \log p(y)=-L(\theta)+\log p(y)\\
  \text{where}\\
   L(\theta)& \overset{\Delta}{=}& \int_x p_{\theta}( x ) \log \left(
  \frac{p(y|x)p(x)}{p_\theta(x)} \right)
\end{eqnarray}

Since the KL divergence is nonnegative, the reward function $L(\theta)$ is a lower bound on the partition function $\log p(y)=\log \int_xp(y|x)p(x)$; the approximation error is therefore
minimized by maximizing the lower bound.

Different choices of $p_{\theta}( x )$ result in different kinds of
approximations.  The popular mean-field approximation decomposes
$p_\theta(x)$ into a product of marginals as $p_\theta(x) =
\prod_{t=1}^T p_\theta( x_t | \theta_t )$, where every random choice
ignores the history $h_t$ of the generative process.

\subsection{Stochastic Gradient Optimization}

Minimizing Eq.~\ref{eq:kl} is typically done by computing derivatives
analytically, setting them equal to zero, solving for a coupled set of
nonlinear equations, and deriving an iterative coordinate descent
algorithm.  However, this approach only works for conjugate
distributions, and fails for highly structured distributions (such as
those represented by, for example, probabilistic programs) that are
not analytically tractable.

One generic approach to solving this is (stochastic) gradient descent
on $L(\theta)$. We estimate the gradient according to the following computation:
\begin{align}
\label{eq:grad}
-\nabla_\theta\: L(\theta)&= \int_x \nabla_\theta \left(p_{\theta}( x ) \log \left(
   \frac {p_\theta(x)} {p(y|x)p(x)}\right) \right)\\
   &= \int_x \nabla_\theta p_\theta(x) \left(\log\left(\frac{p_\theta(x)}{p(y|x) p(x)}\right)\right) + \int_x p_\theta(x) \left(\nabla_\theta \log(p_\theta(x))\right)\\
&= \int_x \nabla_\theta p_\theta(x) \left(\log\left(\frac{p_\theta(x)}{p(y|x) p(x)}\right)\right)\label{eq1}\\
&= \int_x p_\theta(x) \nabla_\theta \log(p_\theta(x)) \left(\log\left(\frac{p_\theta(x)}{p(y|x) p(x)}\right)\right)\label{eq2}\\
&= \int_x p_\theta(x) \nabla_\theta \log(p_\theta(x)) \left(\log\left(\frac{p_\theta(x)}{p(y|x) p(x)}\right)+K \right)\label{eq3}\\
& \approx \frac{1}{N}  \sum_{x^j} \nabla_{\theta} \log p_{\theta}( x^j
)\left(  \log \left( \frac{ p_{\theta}( x^j ) }{ p(y|x^j)p(x^j)} \right)+K \right)\label{eq4}
\end{align}
with $x^j \sim p_{\theta}( x )$, $j=1\ldots N$ and $K$ is an arbitrary constant. To obtain equations (7-9), we repeatedly use the fact that $\nabla \log p_\theta(x)= \frac{\nabla_\theta p_\theta(x)}{p_\theta(x)}$. Furthermore, for equations \eqref{eq1} and \eqref{eq3}, we also use  $\int_x p_\theta(x) \nabla \log p_\theta(x)=0$, since $\int_x p_\theta(x) \nabla \log p_\theta(x)  =\int_x \nabla_\theta p_\theta(x)= \nabla_\theta \int_x p_\theta(x) =\nabla_\theta 1=0$. The purpose of adding the constant $K$ is that it is possible to approximately estimate a value of $K$ (optimal baseline), such that the variance of the Monte-Carlo estimate~\eqref{eq4} of expression~\eqref{eq3} is minimized. As we will see, choosing an appropriate value of $K$ will have drastic effects on the quality of the gradient estimate.

\subsection{Compositional Variational Inference}

Consider a distribution $p(x)$ induced by an arbitrary, unconditional
probabilistic program. Our goal is to estimate marginals of the
conditional distribution $p(x|y)$, which we will call the \emph{target
  program}.  We introduce a variational distribution $p_{\theta}( x
)$, which is defined through another probabilistic program, called the
\emph{variational program}. This distribution is unconditional, so
sampling from it is as easy as running it.

We derive the variational program from the target program. An easy way
to do this is to use a \emph{partial mean-field approximation}: the
target probabilistic program is run forward, and each time an ERP
$x_t$ is encountered, a variational parameter is used in place of
whatever parameters would ordinarily be passed to the ERP.  That is,
instead of sampling $x_t$ from $p_t (x_t\:|\:\psi_t(h_t))$ as in
Eq.~\ref{eq:pplik}, we instead sample from $p_t
(x_t\:|\:\theta_t(h_t))$, where $\theta_t(h_t)$ is an auxiliary
variational parameter (and the true parameter $\psi_t(h_t)$ is
ignored).

Fig.~\ref{fig:ppvp} illustrates this with pseudocode for a
probabilistic program and its variational equivalent: upon
encountering the \texttt{normal} ERP on line 4, instead of using
parameter \texttt{mu}, the variational parameter $\theta_3$ is used
instead (\texttt{normal} is a Gaussian ERP which takes an optional
argument for the mean, and \texttt{rand(a,b)} is uniform over the set
$[a,b]$, with $[0,1]$ as the default argument). Note that a dependency
between $X$ and $M$ exists through the control logic, but not the
parameterization.  Thus, in general, stochastic dependencies due to
the parameters of a variable depending on the outcome of another
variable disappear, but dependencies due to control logic remain
(hence the term \emph{partial mean-field approximation}).


\begin{figure}
\hrule height2pt \vskip 0.04in
\textbf{Probabilistic program A}
\vskip 0.04in \hrule height1pt \vskip 0.04in
\begin{algorithmic}[1]
\STATE \texttt{ M = normal(); }
\STATE \texttt{ if M$>$1 }
\STATE \texttt{ \Sp mu = complex\_deterministic\_func( M ); }
\STATE \texttt{ \Sp X = normal( mu ); }
\STATE \texttt{ else }
\STATE \texttt{ \Sp X = rand(); }
\STATE \texttt{ end; }
\end{algorithmic}
\hrule height1pt \vskip 0.1in

\hrule height2pt \vskip 0.04in
\textbf{Mean-Field variational program A}
\vskip 0.04in \hrule height1pt \vskip 0.04in
\begin{algorithmic}[1]
\STATE \texttt{ M = normal( $\theta_1$ ); }
\STATE \texttt{ if M$>$1 }
\STATE \texttt{ \Sp mu = complex\_deterministic\_func( M ); }
\STATE \texttt{ \Sp X = normal( $\theta_3$ ); }
\STATE \texttt{ else }
\STATE \texttt{ \Sp X = rand($\theta_4,\theta_5$); }
\STATE \texttt{ end; }
\end{algorithmic}
\hrule height1pt \vskip 0.04in
\vskip -0.1in
\caption{A probabilistic program and corresponding variational program}
\label{fig:ppvp}
\end{figure}

This idea can be extended to automatically compute the stochastic
gradient of the variational distribution: we run the forward target
program normally, and whenever a call to an ERP $x_t$ is made, we:
\begin{list}{$\bullet$}{
\leftmargin 0.2in
\listparindent 0in
\itemindent 0in
\topsep 0in
\itemsep 0in
}
\item Sample $x_t$ according to $p_{\theta_t}(x_t)$ (if this is the
  first time the ERP is encountered, initialize $\theta_t$ to an
  arbitrary value, for instance that given by $\psi_t(h_t)$).
\item Compute the log-likelihood $\log p_{\theta_t}(x_t)$ of $x_t$
  according to the mean-field distribution.
\item Compute the log-likelihood $\log p(x_t|h_t)$ of $x_t$ according
  to the target program.
\item Compute the reward $R_t=\log p(x_t|h_t)-\log p_\theta(x_t) $
\item Compute the local gradient $\psi_t= \nabla_{\theta_t} \log p_{\theta_t}(x_t)$.
\end{list}

When the program terminates, we simply compute $\log p(y|x)$, then compute
the gain $R=\sum R_t + \log p(y|x)+K$. The gradient estimate for the
$t^{\text{th}}$ ERP is given by $R \psi_t$, and can be averaged
over many sample traces $x$ for a more accurate estimate.

Thus, the only requirement on the probabilistic program is to be able
to compute the log likelihood of an ERP value, as well as its gradient
with respect to its parameters.  Let us highlight that being able to compute the gradient of the log-likelihood with respect to natural parameters is the only additional requirement compared to an MCMC sampler.

Note that everything above holds: 1)
regardless of conjugacy of distributions in the stochastic program; 2)
regardless of the control logic of the stochastic program; and 3)
regardless of the actual parametrization of $p(x_t;\theta_t)$.  In
particular, we again emphasize that we do not need the gradients of
deterministic structures (for example, the function
\texttt{complex\_deterministic\_func} in Fig.~\ref{fig:ppvp}).

\subsection{Extensions}

Here, we discuss three extensions of our core ideas.

\textbf{Learning inference transfer.}\ \ Assume we wish to run
variational inference for $N$ distinct datasets $y^1,\ldots,y^N$.
Ideally, one should solve a distinct inference problem for each,
yielding distinct $\theta^1,\cdots,\theta^N$. Unfortunately, finding
$\theta^1,\cdots,\theta^N$ does not help to find $\theta^{N+1}$ for a
new dataset $y_{N+1}$.  But perhaps our approach can be used to learn
`approximate samplers': instead of $\theta$ depending on $y$
implicitly via the optimization algorithm, suppose instead that
$\theta_t$ depends on $y$ through some fixed functional form. For
instance, we can assume $\theta_t(y)=\sum_j \alpha_{i,j} f_j(y)$,
where $f_j$ is a known function, then find parameters $\theta_{t,j}$
such that for most observations $y$, the variational distribution
$p_{\theta(y)}(x)$ is a decent approximate sampler to $p(x|y)$.
Gradient estimates of $\alpha$ can be derived similarly to
Eq.~\ref{eq:kl} for arbitrary probabilistic programs.

\textbf{Structured mean-field approximations.}\ \ It is sometimes the
case that a vanilla mean-field distribution is a poor approximation of
the posterior, in which case more structured approximations should be
used~\cite{bouchard2009optimization,ghahramani1997structured,bishop2003structured,geiger2005structured}.
Deriving the variational update for structured mean-field is harder
than vanilla mean-field; however, from a probabilistic program point
of view, a structured mean-field approximation is simply a more
complex (but still unconditional) variational program that could be
derived via program analysis (or perhaps online via RL state-space
estimation), with gradients computed as in the mean-field case.

\textbf{Online probabilistic programming}\ \ One advantage of
stochastic gradients in probabilistic programs is simple
parallelizability.  This can also be done in an online fashion, in a similar fashion to recent work for stochastic variational inference by Blei et al.\ ~\citep{hoffman2010online,wangonline,blei2011stochastic}. Suppose that the set of variables and observations can be separated into a main set $X$,
and a large number $N$ of independent sets of latent variables $X_i$
and observations $Y_i$ (where the $(X_i,Y_i)$ are only allowed to depend on $X$).  For instance, for LDA, $X$ represents the topic
distributions, while the $X_i$ represent the document distribution over
topics and $Y_i$ topic $i$. Recall the gradient for the variational
parameters of $X$ is given by $K \psi_X$, with $K=R_X+ \sum_i
(R_i+\log P(Y_i | X_i,X)$, where $R_X$ is the sum of rewards for all
ERPs in $X$, and $R_i$ is the sum of rewards for all ERPs in
$X_i$. $K$ can be rewritten as $X + N E[R_v + \log P(Y_v | X_v,X)$,
where $v$ is a random integer in $\{1,\ldots,N\}$. The expectation can
be approximately computed in an online fashion, allowing the update of the
estimate of $X$ without manipulating the entire data set $Y$.

\section{Experiments: LDA and QMR}

\begin{figure}[!htb]\label{fig:enac_results}
\subfigure[Results on QMR]{
\includegraphics[width=.31\textwidth]{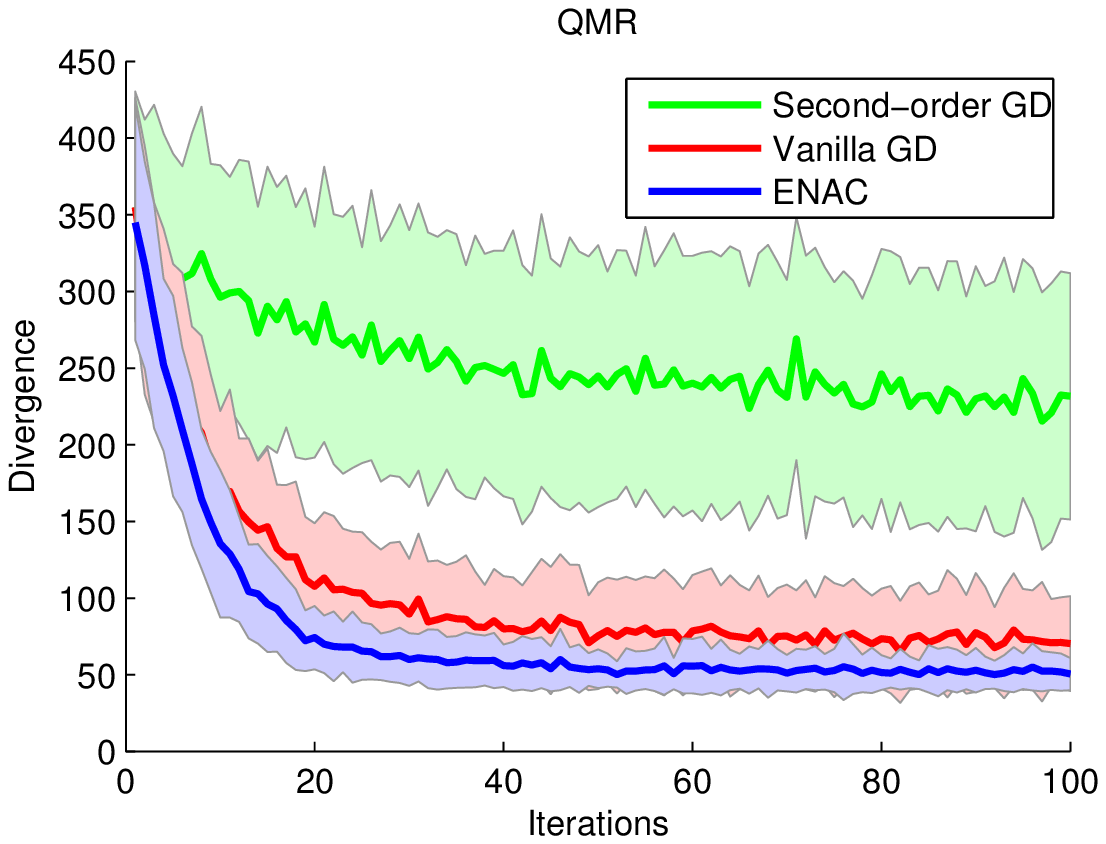}
}
\subfigure[Results on LDA]{
\includegraphics[width=.31\textwidth]{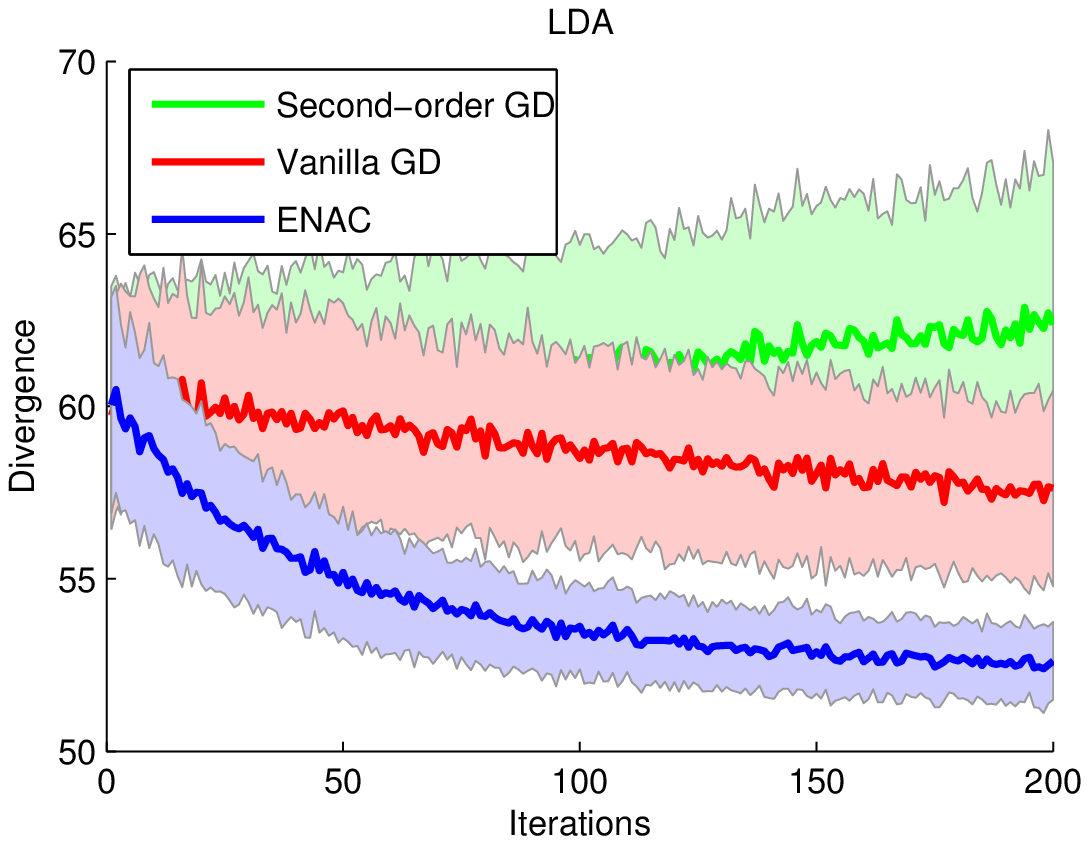}
}
\subfigure[Steepest descent vs. conjugate gradients]{
\includegraphics[width=.31\textwidth]{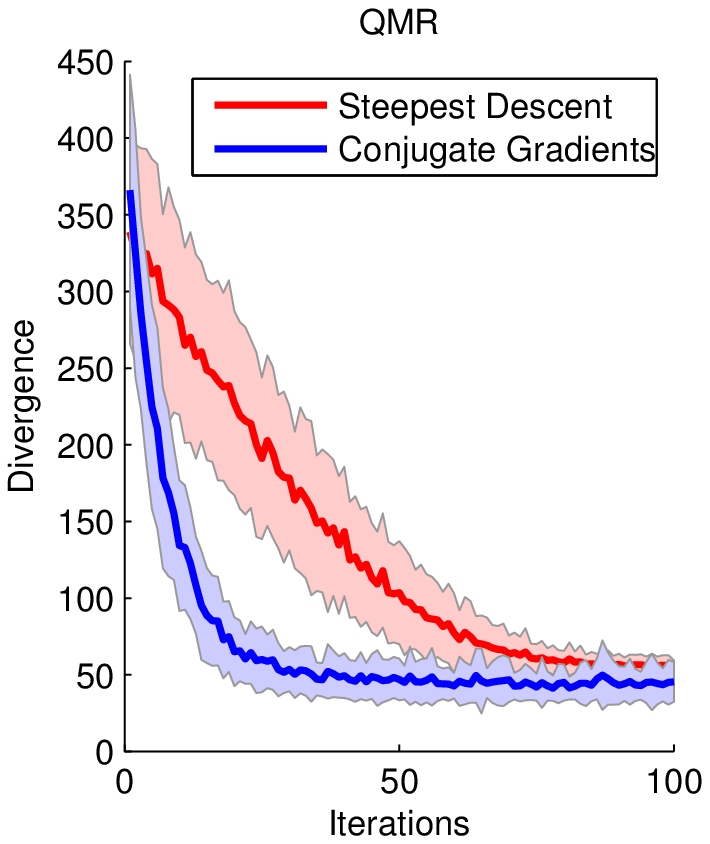}
}
\caption{Numerical simulations of AVI}

\end{figure}

We tested automated variational inference on two common inference benchmarks: the
QMR-DT network (a binary bipartite graphical model with noisy-or
directed links) and LDA (a popular topic model).  We compared three
algorithms: 
\begin{list}{$\bullet$}{
\leftmargin 0.2in
\listparindent 0in
\itemindent 0in
\topsep 0in
\itemsep 0in
}
\item The first is vanilla stochastic gradient descent on
  Eq.~\ref{eq:kl}, with the gradients given by Eq.~\ref{eq:grad}.
\item The Episodic Natural Actor Critic algorithm, a version of the algorithm connecting variational inference to reinforcement learning -- details are reserved for a longer version of this paper. An important feature of ENAC is optimizing over the baseline constant $K$.
\item A second-order gradient descent (SOGD) algorithm which estimates
  the Fisher information matrix $F_\theta$ in the same way as the ENAC
  algorithm, and uses it as curvature information.
\end{list}
For each algorithm, we set $M=10$ (i.e., far fewer roll-outs than
parameters).  All three algorithms were given the same ``budget'' of
samples; they used them in different ways.  All three algorithms
estimated a gradient $\hat{g}(\theta)$; these were used in a steepest
descent optimizer: $\theta = \theta + \alpha \hat{g}(\theta)$ with
stepsize $\alpha$.  All three algorithms used the same stepsize; in
addition, the gradients $g$ were scaled to have unit norm.  The
experiment thus directly compares the quality of the direction of the
gradient estimate.

Fig.~\ref{fig:enac_results} shows the results.  The ENAC algorithm
shows faster convergence and lower variance than steepest descent,
while SOGD fares poorly (and even diverges in the case of LDA).
Fig.~\ref{fig:enac_results} also shows that the gradients from ENAC
can be used either with steepest descent or a conjugate gradient
optimizer; conjugate gradients converge faster.

Because both SOGD and ENAC estimate $F_\theta$ in the same way, we
conclude that the performance advantage of ENAC is \emph{not} due
solely to its use of second-order information: the additional step of
estimating the baseline improves performance significantly.

Once converged, the estimated variational program allows very fast approximate sampling from the posterior, at a fraction of the cost of a sample obtained using MCMC sampling. Samples from the variational program can also be used as warm starts for MCMC sampling. 

\section{Related Work}

\nocite{herbst10,wainwright2008graphical,beal2003variational,winn2006variational,peters05,honkela2008natural,amari98,kakade02,welling2008hybrid}

Natural conjugate gradients for variational inference are investigated
in~\citep{honkela2008natural}, but the analysis is mostly devoted to
the case where the variational approximation is Gaussian, and the
resulting gradient equation involves an integral which is not
necessarily tractable.

The use of variational inference in probabilistic programming is
explored in~\cite{harik2010variational}. The authors similarly note
that it is easy to sample from the variational program. However, they
only use this observation to estimate the free energy of the
variational program, but they do not estimate the gradient of that
free energy. While they do highlight the need for optimizing the
parameters of the variational program, they do not offer a general
algorithm for doing so, instead suggesting rejection sampling or
importance sampling.

Use of stochastic approximations for variational inference is also
used by Carbonetto~\cite{carbonetto2009stochastic}. Their approach is
very different from ours: they use Sequential Monte Carlo to refine
gradient estimates, and require that the family of variational
distributions contains the target distribution. While their approach
is fairly general, it cannot be automatically generated for
arbitrarily complex probabilistic models.

Finally, stochastic gradient methods are also used in online variational
inference algorithms, in particular in the work of Blei et al.\ in stochastic variational inference (for instance, online LDA~\cite{hoffman2010online}, online HDP~\cite{wangonline}, and more generally under conjugacy assumptions~\cite{blei2011stochastic}), as a way to refine estimates of latent variable distributions without processing all the observations. However, this approach requires a manual derivation of the variational equation for coordinate descent, which is only possible under conjugacy assumptions which will in general not hold for arbitrary probabilistic programs.

\bibliographystyle{unsrtnat}
\bibliography{avi}

\begin{thebibliography}{30}
\providecommand{\natexlab}[1]{#1}
\providecommand{\url}[1]{\texttt{#1}}
\expandafter\ifx\csname urlstyle\endcsname\relax
  \providecommand{\doi}[1]{doi: #1}\else
  \providecommand{\doi}{doi: \begingroup \urlstyle{rm}\Url}\fi

\bibitem[Milch et~al.(2005)Milch, Marthi, Russell, Sontag, Ong, and
  Kolobov]{milch05}
Brian Milch, Bhaskara Marthi, Stuart Russell, David Sontag, Daniel~L. Ong, and
  Andrey Kolobov.
\newblock {BLOG}: Probabilistic models with unknown objects.
\newblock In \emph{International Joint Conference on Artificial Intelligence
  ({IJCAI})}, pages 1352--1359, 2005.

\bibitem[Sato and Kameya(1997)]{sato97}
T.~Sato and Y.~Kameya.
\newblock {PRISM}: A symbolic-statistical modeling language.
\newblock In \emph{International Joint Conference on Artificial Intelligence
  ({IJCAI})}, 1997.

\bibitem[Kersting and Raedt(2007)]{kersting07}
K.~Kersting and L.~De Raedt.
\newblock Bayesian logic programming: Theory and tool.
\newblock In L.~Getoor and B.~Taskar, editors, \emph{An Introduction to
  Statistical Relational Learning}. MIT Press, 2007.

\bibitem[Muggleton(1996)]{Muggleton96stochasticlogic}
Stephen Muggleton.
\newblock Stochastic logic programs.
\newblock In \emph{New Generation Computing}. Academic Press, 1996.

\bibitem[Poole(2008)]{poole08}
David Poole.
\newblock The independent choice logic and beyond.
\newblock pages 222--243, 2008.

\bibitem[Pfeffer(2001)]{pfeffer01}
Avi Pfeffer.
\newblock {IBAL}: A probabilistic rational programming language.
\newblock In \emph{International Joint Conference on Artificial Intelligence
  ({IJCAI})}, pages 733--740. Morgan Kaufmann Publ., 2001.

\bibitem[Radul(2007)]{radul07}
Alexhey Radul.
\newblock Report on the probabilistic language scheme.
\newblock Technical Report MIT-CSAIL-TR-2007-059, Massachusetts Institute of
  Technology, 2007.

\bibitem[Park et~al.(2008)Park, Pfenning, and Thrun]{park08}
Sungwoo Park, Frank Pfenning, and Sebastian Thrun.
\newblock A probabilistic language based on sampling functions.
\newblock \emph{ACM Trans. Program. Lang. Syst.}, 31\penalty0 (1):\penalty0
  1--46, 2008.

\bibitem[Goodman et~al.(2008)Goodman, Mansinghka, Roy, Bonawitz, and
  Tenenbaum]{goodman08}
Noah Goodman, Vikash Mansinghka, Daniel Roy, Keith Bonawitz, and Joshua
  Tenenbaum.
\newblock Church: a language for generative models.
\newblock In \emph{Uncertainty in Artificial Intelligence ({UAI})}, 2008.

\bibitem[Wingate et~al.(2011)Wingate, Stuhlmueller, and Goodman]{wingate11}
David Wingate, Andreas Stuhlmueller, and Noah~D. Goodman.
\newblock Lightweight implementations of probabilistic programming languages
  via transformational compilation.
\newblock In \emph{International Conference on Artificial Intelligence and
  Statistics ({AISTATS})}, 2011.

\bibitem[Kiselyov and Shan(2009)]{kiselyovS09}
Oleg Kiselyov and {Chung-chieh} Shan.
\newblock Embedded probabilistic programming.
\newblock In \emph{Domain-Specific Languages}, pages 360--384, 2009.

\bibitem[Beal and MA(2003)]{beal2003variational}
M.J. Beal and M.~MA.
\newblock Variational algorithms for approximate bayesian inference.
\newblock \emph{Unpublished doctoral dissertation, University College London},
  2003.

\bibitem[Jordan et~al.(1999)Jordan, Ghahramani, and Jaakola]{jordan99}
Michael~I. Jordan, Zoubin Ghahramani, and Tommi Jaakola.
\newblock Introduction to variational methods for graphical models.
\newblock \emph{Machine Learning}, \penalty0 (37):\penalty0 183--233, 1999.

\bibitem[Winn and Bishop(2006)]{winn2006variational}
J.~Winn and C.M. Bishop.
\newblock Variational message passing.
\newblock \emph{Journal of Machine Learning Research}, 6\penalty0 (1):\penalty0
  661, 2006.

\bibitem[Bouchard-C{\^o}t{\'e} and Jordan(2009)]{bouchard2009optimization}
A.~Bouchard-C{\^o}t{\'e} and M.I. Jordan.
\newblock Optimization of structured mean field objectives.
\newblock In \emph{Proceedings of the Twenty-Fifth Conference on Uncertainty in
  Artificial Intelligence}, pages 67--74. AUAI Press, 2009.

\bibitem[Ghahramani(1997)]{ghahramani1997structured}
Z.~Ghahramani.
\newblock On structured variational approximations.
\newblock \emph{University of Toronto Technical Report, CRG-TR-97-1}, 1997.

\bibitem[Bishop and Winn(2003)]{bishop2003structured}
C.M. Bishop and J.~Winn.
\newblock Structured variational distributions in vibes.
\newblock \emph{Proceedings Artificial Intelligence and Statistics, Key West,
  Florida}, 2003.

\bibitem[Geiger and Meek(2005)]{geiger2005structured}
D.~Geiger and C.~Meek.
\newblock Structured variational inference procedures and their realizations.
\newblock In \emph{Proc. AIStats}. Citeseer, 2005.

\bibitem[Hoffman et~al.(2010)Hoffman, Blei, and Bach]{hoffman2010online}
M.D. Hoffman, D.M. Blei, and F.~Bach.
\newblock Online learning for latent dirichlet allocation.
\newblock \emph{Advances in Neural Information Processing Systems},
  23:\penalty0 856--864, 2010.

\bibitem[Wang et~al.()Wang, Paisley, and Blei]{wangonline}
C.~Wang, J.~Paisley, and D.M. Blei.
\newblock Online variational inference for the hierarchical dirichlet process.

\bibitem[Blei(2011)]{blei2011stochastic}
D.M. Blei.
\newblock Stochastic variational inference.
\newblock 2011.

\bibitem[Herbst(2010)]{herbst10}
Edward Herbst.
\newblock Gradient and {H}essian-based {MCMC} for {DSGE} models (job market
  paper), 2010.

\bibitem[Wainwright and Jordan(2008)]{wainwright2008graphical}
M.J. Wainwright and M.I. Jordan.
\newblock Graphical models, exponential families, and variational inference.
\newblock \emph{Foundations and Trends{\textregistered} in Machine Learning},
  1\penalty0 (1-2):\penalty0 1--305, 2008.

\bibitem[Peters et~al.(2005)Peters, Vijayakumar, and Schaal]{peters05}
Jan Peters, Sethu Vijayakumar, and Stefan Schaal.
\newblock Natural actor-critic.
\newblock In \emph{European Conference on Machine Learning (ECML)}, pages
  280--291, 2005.

\bibitem[Honkela et~al.(2008)Honkela, Tornio, Raiko, and
  Karhunen]{honkela2008natural}
A.~Honkela, M.~Tornio, T.~Raiko, and J.~Karhunen.
\newblock Natural conjugate gradient in variational inference.
\newblock In \emph{Neural Information Processing}, pages 305--314. Springer,
  2008.

\bibitem[Amari(1998)]{amari98}
S.~Amari.
\newblock Natural gradient works efficiently in learning.
\newblock \emph{Neural Computation}, \penalty0 (10), 1998.

\bibitem[Kakade(2002)]{kakade02}
Sham~A. Kakade.
\newblock Natural policy gradient.
\newblock In \emph{Neural Information Processing Systems (NIPS)}, 2002.

\bibitem[Welling et~al.(2008)Welling, Teh, and Kappen]{welling2008hybrid}
M.~Welling, Y.W. Teh, and B.~Kappen.
\newblock Hybrid variational/gibbs collapsed inference in topic models.
\newblock In \emph{Proceedings of the Conference on Uncertainty in Artifical
  Intelligence (UAI)}, pages 587--594. Citeseer, 2008.

\bibitem[Harik and Shazeer(2010)]{harik2010variational}
G.~Harik and N.~Shazeer.
\newblock Variational program inference.
\newblock \emph{Arxiv preprint arXiv:1006.0991}, 2010.

\bibitem[Carbonetto et~al.(2009)Carbonetto, King, and
  Hamze]{carbonetto2009stochastic}
P.~Carbonetto, M.~King, and F.~Hamze.
\newblock A stochastic approximation method for inference in probabilistic
  graphical models.
\newblock In \emph{NIPS}, volume~22, pages 216--224. Citeseer, 2009.

\end{thebibliography}

\end{document}